# Survey of Nearest Neighbor Techniques

Nitin Bhatia (Corres. Author)
Department of Computer Science
DAV College
Jalandhar, INDIA
n_bhatia78@yahoo.com

Vandana
SSCS
Deputy Commissioner's Office
Jalandhar, INDIA
vandana_ashev@yahoo.co.in

*Abstract*— **The nearest neighbor (NN) technique is very simple, highly efficient and effective in the field of pattern recognition, text categorization, object recognition etc. Its simplicity is its main advantage, but the disadvantages can't be ignored even. The memory requirement and computation complexity also matter. Many techniques are developed to overcome these limitations. NN techniques are broadly classified into structure less and structure based techniques. In this paper, we present the survey of such techniques. Weighted kNN, Model based kNN, Condensed NN, Reduced NN, Generalized NN are structure less techniques whereas k-d tree, ball tree, Principal Axis Tree, Nearest Feature Line, Tunable NN, Orthogonal Search Tree are structure based algorithms developed on the basis of kNN. The structure less method overcome memory limitation and structure based techniques reduce the computational complexity.**

## I. INTRODUCTION

The nearest neighbor (NN) rule identifies the category of unknown data point on the basis of its nearest neighbor whose class is already known. This rule is widely used in pattern recognition [13, 14], text categorization [15-17], ranking models [18], object recognition [20] and event recognition [19] applications.

T. M. Cover and P. E. Hart purpose k-nearest neighbor (kNN) in which nearest neighbor is calculated on the basis of value of k, that specifies how many nearest neighbors are to be considered to define class of a sample data point [1]. T. Bailey and A. K. Jain improve kNN which is based on weights [2]. The training points are assigned weights according to their distances from sample data point. But still, the computational complexity and memory requirements remain the main concern always. To overcome memory limitation, size of data set is reduced. For this, the repeated patterns, which do not add extra information, are eliminated from training samples [3-5]. To further improve, the data points which do not affect the result are also eliminated from training data set [6]. Besides the time and memory limitation, another point which should be taken care of, is the value of k, on the basis of which category of the unknown sample is determined. Gongde Guo selects the value of k using model based approach [7]. The model proposed automatically selects the value of k. Similarly, many improvements are proposed to improve speed of classical kNN using concept of ranking [8], false neighbor information [9], clustering [10]. The NN training data set can be structured using various techniques to improve over memory limitation of kNN. The kNN implementation can be done using ball tree [21, 22], k-d tree [23], nearest feature line (NFL) [24], tunable metric [26], principal axis search tree [28] and orthogonal search tree [29]. The tree structured training data is divided into nodes, whereas techniques like NFL and tunable metric divide the training data set according to planes. These algorithms increase the speed of basic kNN algorithm.

## II. NEAREST NEIGHBOR TECHNIQUES

Nearest neighbor techniques are divided into two categories: 1) Structure less and 2) Structure Based.

### A. *Structure less NN techniques*

The k-nearest neighbor lies in first category in which whole data is classified into training data and sample data point. Distance is evaluated from all training points to sample point and the point with lowest distance is called nearest neighbor.

This technique is very easy to implement but value of k affects the result in some cases. Bailey uses weights with classical kNN and gives algorithm named weighted kNN (WkNN) [2]. WkNN evaluates the distances as per value of k and weights are assigned to each calculated value, and then nearest neighbor is decided and class is assigned to sample data point. The Condensed Nearest Neighbor (CNN) algorithm stores the patterns one by one and eliminates the duplicate ones. Hence, CNN removes the data points which do not add more information and show similarity with other training data set. The Reduced Nearest Neighbor (RNN) is improvement over CNN; it includes one more step that is elimination of the patterns which are not affecting the training data set result. The another technique called Model Based kNN selects similarity measures and create a 'similarity matrix' from given training set. Then, in the same category, largest local neighbor is found that covers large number of neighbors and a data tuple is located with largest global neighborhood. These steps are repeated until all data tuples are grouped. Once data is formed using model, kNN is executed to specify category of unknown sample. Subash C. Bagui and Sikha Bagui [8] improve the kNN by introducing the concept of ranks. The method pools all the observations belonging to different categories and assigns rank to each category data in ascending order. Then observations are counted and on the basis of ranks class is assigned to unknown sample. It is very much useful in case of multi-variants data. In Modified kNN, which is modification of WkNN validity of all data samples in the training data set is computed, accordingly weights are assigned and then validity and weight both together set basis for classifying the class of the sample data point. Yong zeng, Yupu Zeng and Liang Zhou define the new concept to classify sample data point. The method introduces the pseudo neighbor, which is not the actual nearest neighbor; but a new nearest neighbor is selected on the basis of value of weighted sum of distances of kNN of unclassified patterns in each class. Then Euclidean distance is evaluated and pseudo neighbor with greater weight is found and classified for unknown sample. In the technique purposed by Zhou Yong [11], Clustering is used to calculate nearest neighbor. The steps include, first of all removing the samples which are lying near to the border, from training set. Cluster each training set by k value clustering and all cluster centers form new training set. Assign weight to each cluster according to number of training samples each cluster have.

### *B. Structure based NN techniques*

The second category of nearest neighbor techniques is based on structures of data like Ball Tree, k-d Tree, principal axis Tree (PAT), orthogonal structure Tree (OST), Nearest feature line (NFL), Center Line (CL) etc. Ting Liu introduces the concept of Ball Tree. A ball tree is a binary tree and constructed using top down approach. This technique is improvement over kNN in terms of speed. The leaves of the tree contain relevant information and internal nodes are used to guide efficient search through leaves. The k-dimensional trees divide the training data into two parts, right node and left node. Left or right side of tree is searched according to query records. After reaching the terminal node, records in terminal node are examined to find the closest data node to query record. The concept of NFL given by Stan Z.Li and Chan K.L. [24] divide the training data into plane. A feature line (FL) is used to find nearest neighbor. For this, FL distance between query point and each pair of feature line is calculated for each class. The resultant is set of distances. The evaluated distances are sorted into ascending order and the NFL distance is assigned as rank 1. An improvement made over NFL is Local Nearest Neighbor which evaluates the feature line and feature point in each class, for points only, whose corresponding prototypes are neighbors of query point. Yongli Zhou and Changshui Zhang introduce [26] new metric for evaluating distances for NFL rather than feature line. This new metric is termed as “Tunable Metric”. It follows the same procedure as NFL but at first stage it uses tunable metric to calculate distance and then implement steps of NFL. Center Based Nearest Neighbor is improvement over NFL and Tunable Nearest Neighbor. It uses center base line (CL) that connects sample point with known labeled points. First of all CL is calculated, which is straight line passing through training sample and center of class. Then distance is evaluated from query point to CL, and nearest neighbor is evaluated. PAT permits to divide the training data into efficient manner in term of speed for nearest neighbor evaluation. It consists of two phases 1) PAT Construction 2) PAT Search. PAT uses principal component analysis (PCA) and divides the data set into regions containing the same number of points. Once tree is formed kNN is used to search nearest neighbor in PAT. The regions can be determined for given point using binary search. The OST uses orthogonal vector. It is an improvement over PAT for speedup the process. It uses concept of “length (norm)”, which is evaluated at first stage. Then orthogonal search tree is formed by creating a root node and assigning all data points to this node. Then left and right nodes are formed using pop operation.

TABLE I. COMPARISON OF NEAREST NEIGHBOR TECHNIQUES

| Sr No | Technique | Key Idea | Advantages | Disadvantages | Target Data |
|---|---|---|---|---|---|
| 1. | k Nearest Neighbor (kNN) [1] | Uses nearest neighbor rule | 1. training is very fast<br>2. Simple and easy to learn<br>3. Robust to noisy training data<br>4.Effective if training data is large | 1. Biased by value of k<br>2.Computation Complexity<br>3.Memory limitation<br>4. Being a supervised learning lazy algorithm i.e. runs slowly<br>5. Easily fooled by irrelevant attributes | large data samples |
| 2. | Weighted k nearest neighbor (WkNN) [2] | Assign weights to neighbors as per distance calculated | 1. Overcomes limitations of kNN of assigning equal weight to k neighbors implicitly.<br>2. Use all training samples not just k.<br>3. Makes the algorithm global one | 1. Computation complexity increases in calculating weights<br>2. Algorithm runs slow | Large sample data |
| 3. | Condensed nearest neighbor (CNN) [3,4,5] | Eliminate data sets which show similarity and do not add extra information | 1. Reduce size of training data<br>2. Improve query time and memory requirements<br>3.Reduce the recognition rate | 1. CNN is order dependent; it is unlikely to pick up points on boundary.<br>2. Computation Complexity | Data set where memory requirement is main concern |
| 4. | Reduced Nearest Neigh (RNN) [6] | Remove patterns which do not affect the training data set results | 1. Reduce size of training data and eliminate templates<br>2. Improve query time and memory requirements<br>3.Reduce the recognition rate | 1.Computational Complexity<br>2.Cost is high<br>3.Time Consuming | Large data set |
| 5. | Model based k nearest neighbor (MkNN) [7] | Model is constructed from data and classify new data using model | 1. More classification accuracy<br>2.Value of k is selected automatically<br>3.High efficiency as reduce number of data points | 1.Do not consider marginal data outside the region | Dynamic web mining for large repository |
| 6. | Rank nearest neighbor (kRNN) [8] | Assign ranks to training data for each category | 1.Performs better when there are too much variations between features<br>2.Robust as based on rank | 1.Multivariate kRNN depends on distribution of the data | Class distribution of Gaussian nature |
| | | | 3.Less computation complexity as compare to kNN | | |
| 7. | Modified k nearest neighbor (MkNN) [10] | Uses weights and validity of data point to classify nearest neighbor | 1.Partially overcome low accuracy of WkNN<br>2.Stable and robust | 1.Computation Complexity | Methods facing outlets |
| 8. | Pseudo/Generalized Nearest Neighbor (GNN) [9] | Utilizes information of n-1 neighbors also instead of only nearest neighbor | 1.uses n-1 classes which consider the whole training data set | 1.does not hold good for small data<br>2.Computationa;l complexity | Large data set |
| 9. | Clustered k nearest neighbor [11] | Clusters are formed to select nearest neighbor | 1.Overcome defect of uneven distributions of training samples<br>2.Robust in nature | 1.Selection of threshold parameter is difficult before running algorithm<br>2.Biased by value of k for clustering | Text Classification |
| 10. | Ball Tree k nearest neighbor (KNS1) [21,22] | Uses ball tree structure to improve kNN speed | 1.Tune well to structure of represented data<br>2.Deal well with high dimensional entities<br>3.Easy to implement | 1.Costly insertion algorithms<br>2.As distance increases KNS1 degrades | Geometric Learning tasks like robotic, vision, speech, graphics |
| 11. | k-d tree nearest neighbor (kdNN) [23] | divide the training data exactly into half plane | 1.Produce perfectly balanced tree<br>2.Fast and simple | 1.More computation<br>2.Require intensive search<br>3.Blindly slice points into half which may miss data structure | organization of multi dimensional points |
| 12. | Nearest feature Line Neighbor (NFL) [24] | take advantage of multiple templates per class | 1.Improve classification accuracy<br>2.Highly effective for small size<br>3.utilises information ignored in nearest neighbor i.e. templates per class | 1.Fail when prototype in NFL is far away from query point<br>2.Computations Complexity<br>3.To describe features points by straight line is hard task | Face Recognition problems |
| 13. | Local Nearest Neighbor [25] | Focus on nearest neighbor prototype of query point | 1.Cover limitations of NFL | 1.Number of Computations | Face Recognition |
| 14. | Tunable Nearest Neighbor (TNN) [26] | A tunable metric is used | 1.Effective for small data sets | 1.Large number of computations | Discrimination problems |
| 15. | Center based Nearest Neighbor (CNN) [27] | A Center Line is calculated | 1.Highly efficient for small data sets | 1. Large number of computations | Pattern Recognition |
| 16. | Principal Axis Tree Nearest Neighbor (PAT) [28] | Uses PAT | 1.Good performance<br>2.Fast Search | 1.Computation Time | Pattern Recognition |
| 17. | Orthogonal Search Tree Nearest Neighbor [29] | Uses Orthogonal Trees | 1.Less Computation time<br>2.Effective for large data sets | 1.Query time is more | Pattern Recognition |

## III. CONCLUSION

We compared the nearest neighbor techniques. Some of them are structure less and some are structured base. Both kinds of techniques are improvements over basic kNN techniques. Improvements are proposed by researchers to gain speed efficiency as well as space efficiency. Every technique hold good in particular field under particular circumstances.